\documentclass{bmvc2k}

\usepackage[utf8]{inputenc}
\usepackage{epstopdf}
\usepackage{amsmath}
\usepackage{amssymb}
\usepackage{bm}
\usepackage{floatrow}
\newfloatcommand{capbtabbox}{table}[][\FBwidth]

\title{Camera Pose Estimation from Lines using Pl\"{u}cker Coordinates}

\addauthor{Bronislav Přibyl}{ipribyl@fit.vutbr.cz}{1}
\addauthor{Pavel Zemčík}{zemcik@fit.vutbr.cz}{1}
\addauthor{Martin Čadík}{cadik@fit.vutbr.cz}{1}

\addinstitution{{\fontsize{9.5pt}{1.1em}\selectfont
 \hspace{-3.7em}Graph@FIT, CPhoto@FIT\\
 \hspace{-3.7em}Department of Computer Graphics and Multimedia\\
 \hspace{-3.7em}Faculty of Information Technology\\
 \hspace{-3.7em}Brno University of Technology\\
 \hspace{-3.7em}Božetěchova 2\\
 \hspace{-3.7em}612\,66 Brno\\
 \hspace{-3.7em}Czech Republic
 }
}
\def\eg{\emph{e.g}\bmvaOneDot}

\def\etal{\emph{et al}\bmvaOneDot}

\runninghead{Přibyl \etal}{Camera Pose Estimation from Lines using Pl\"{u}cker Coords.}

\begin{document}

\maketitle

\begin{abstract}
Correspondences between 3D lines and their 2D images captured by a camera are often used to determine position and orientation of the camera in space. In this work, we propose a novel algebraic algorithm to estimate the camera pose. We parameterize 3D lines using Pl\"{u}cker coordinates that allow linear projection of the lines into the image. A line projection matrix is estimated using Linear Least Squares and the camera pose is then extracted from the matrix. An algebraic approach to handle mismatched line correspondences is also included. The proposed algorithm is an order of magnitude faster yet comparably accurate and robust to the state-of-the-art, it does not require initialization, and it yields only one solution. The described method requires at least 9 lines and is particularly suitable for scenarios with 25 and more lines, as also shown in the results.
\end{abstract}

\section{Introduction}

Camera pose estimation is the task of determining the position and orientation of a camera in 3D space and it has many applications in computer vision, cartography, and related fields. Augmented reality, robot localization, navigation, or 3D reconstruction are just a few of them. To estimate the camera pose, correspondences between known real world features and their counterparts in the image plane of the camera have to be learned. The features can be \eg points, lines, or combinations of both~\cite{kuang2013pose}. The task has been solved using \emph{point correspondences} first~\cite{lowe1987three,fischler1981random}. This is called the \emph{Perspective-n-Point} (PnP) problem and it still enjoys attention of the scientific community~\cite{ferraz2014very}. Camera pose can also be estimated using \emph{line correspondences}, which is called the \emph{Perspective-n-Line} (PnL) problem. A remarkable progress in solving PnL has been achieved in the last years~\cite{mirzaei2011globally,zhang2013robust,bhat2014line}, particularly thanks to the work of Mirzaei and Roumeliotis~\cite{mirzaei2011globally} and more recently to the work of Zhang \etal\cite{zhang2013robust}. Both of the methods are accurate, cope well with noisy data, and they are more efficient than the previously known methods. Computational efficiency  is a critical aspect for many applications  and we show that it can be pushed even further.

We propose an efficient solution to the PnL problem which is substantially faster yet accurate and robust compared to the state-of-the-art \cite{mirzaei2011globally,zhang2013robust}. The idea is to parameterize the 3D lines using Pl\"{u}cker coordinates~\cite{bartoli2005structure} to allow using Linear Least Squares to estimate the projection matrix. The camera pose parameters are then extracted from the projection matrix by posterior constraint enforcement.

The proposed method (\textbf{i}) is more than the order of magnitude faster than the state-of-the-art \cite{mirzaei2011globally,zhang2013robust}, (\textbf{ii}) yields only one solution of the PnL problem, and (\textbf{iii}) similarly to the state-of-the-art, copes well with image noise, and is initialization-free. These advantages make the proposed method particularly suitable for scenarios with many lines. The method needs 9 lines in the minimal case, so it is not practical for a RANSAC-like framework because it would result in increased number of iterations. To eliminate this, we involve an alternative algebraic scheme to deal with mismatched line correspondences.

The rest of this paper is organized as follows. We present a review of related work in Section~\ref{sec:related}. Then we state the basics of parameterizing 3D lines using Pl\"{u}cker coordinates in Section~\ref{sec:pose-estim}, show how the lines are projected onto the image plane and how we exploit it to estimate the camera pose. We evaluate the performance of our method using simulations and real-world experiments in Section~\ref{sec:evaluation}, and conclude in Section~\ref{sec:conclusions}.

\vspace{-1em}
\section{Related work}
\label{sec:related}
\vspace{-0.5em}

The task of camera pose estimation from line correspondences is receiving attention for more than two decades. Some of the earliest works are the ones of Liu \etal\cite{liu1990determination} and Dhome \etal\cite{dhome1989determination}. They introduce two different ways to deal with the PnL problem which can be tracked until today -- algebraic and iterative approaches.

The \emph{iterative approaches} consider pose estimation as a Nonlinear Least Squares problem by iteratively minimizing specific cost function, which usually has a geometrical meaning. Earlier works \cite{liu1990determination} attempted to estimate the camera position and orientation separately while the latter ones \cite{kumar1994robust, christy1999iterative, david2003simultaneous} favour simultaneous estimation. The problem is that majority of iterative algorithms do not guarantee convergence to the global minimum; therefore, without an accurate initialization, the estimated pose is often far from the true camera pose.

The \emph{algebraic approaches} estimate the camera pose by solving a system of (usually polynomial) equations, minimizing an algebraic error. Dhome \etal\cite{dhome1989determination} and Chen~\cite{chen1990pose} solve the minimal problem of pose estimation from 3 line correspondences whereas Ansar and Daniilidis~\cite{ansar2003linear} work with 4 or more lines. Their algorithm has quadratic computational complexity depending on the number of lines and it may fail if the polynomial system has more than 1 solution. More crucial disadvantage of these methods is that they become unstable in the presence of image noise and must be plugged into a RANSAC or similar loop.

Recently, two major improvements of algebraic approaches have been achieved. First, Mirzaei and Roumeliotis~\cite{mirzaei2011globally} proposed a method which is both efficient (linear computational complexity depending on the number of lines) and robust in the presence of image noise. The cases with 3 or more lines can be handled.
A polynomial system with 27 candidate solutions is constructed and solved through the eigendecomposition of a multiplication matrix.
Camera orientations having the least square error are considered to be the optimal ones. Camera positions are obtained separately using the Linear Least Squares. Nonetheless, the problem of this algorithm is that it often yields multiple solutions.

The second recent improvement is the work of Zhang \etal\cite{zhang2013robust}. Their method works with 4 or more lines and is more accurate and robust than the method of Mirzaei and Roumeliotis. An intermediate model coordinate system is used in the method of Zhang \etal, which is aligned with a 3D line of longest projection. The lines are divided into triples for each of which a P3L polynomial is formed. The optimal solution of the polynomial system is selected from the roots of its derivative in terms of a least squares residual. A drawback of this algorithm is that the computational time increases strongly for higher number of lines.

In this paper, we propose an algebraic solution to the PnL problem which is an order of magnitude faster than the two described state-of-the-art methods yet it is comparably accurate and robust in the presence of image noise.

\vspace{-1em}
\section{Pose estimation using Pl\"{u}cker coordinates}
\label{sec:pose-estim}
\vspace{-0.5em}

Let us assume that we have (\textbf{i}) a calibrated pinhole camera and (\textbf{ii}) correspondences between 3D lines and their images obtained by the camera.
The 3D lines are parameterized using Pl\"{u}cker coordinates (Section~\ref{subsec:plucker}) which allows linear projection of the lines into the image (Section~\ref{subsec:projection}). A line projection matrix can thus be estimated using Linear Least Squares (Section~\ref{subsec:lineprojmat}). The camera pose parameters are extracted from the line projection matrix (Section~\ref{subsec:estimcampose}). An outlier rejection scheme must be employed in cases where line mismatches occur (Section~\ref{subsec:outliers}). For the pseudocode of our algorithm, please refer to Appendix~A in the supplementary material~\cite{pribyl2015supplementary}. An implementation of our algorithm in Matlab is also provided.

Let us now define the coordinate systems: a world coordinate system $\{W\}$ and a camera coordinate system $\{C\}$, both are right-handed. The camera $x$-axis goes right, the $y$-axis goes up and the $z$-axis goes behind the camera, so that the points situated in front of the camera have negative $z$ coordinates in $\{C\}$. A homogeneous 3D point
$\mathbf{A}^{\scriptscriptstyle W} = (a^{\scriptscriptstyle W}_x ~ a^{\scriptscriptstyle W}_y ~ a^{\scriptscriptstyle W}_z ~ a^{\scriptscriptstyle W}_w)^\top$
in $\{W\}$ is transformed into a point
$\mathbf{A}^{\scriptscriptstyle C} = (a^{\scriptscriptstyle C}_x ~ a^{\scriptscriptstyle C}_y ~ a^{\scriptscriptstyle C}_z ~ a^{\scriptscriptstyle C}_w)^\top$
in $\{C\}$ as

\begin{equation}
  \mathbf{A}^{\scriptscriptstyle C} ~=~
	\left( \begin{array}{lc}
		\mathbf{R} & -\mathbf{R} \mathbf{t} \\
		\mathbf{0}_{\scriptscriptstyle 1 \times 3} & 1 
	\end{array} \right) ~
	\mathbf{A}^{\scriptscriptstyle W}~,
  \label{eq:transform}
\end{equation}

\noindent where $\mathbf{R}$ is a $3 \times 3$ rotation matrix describing the orientation of the camera in $\{W\}$ by means of three consecutive rotations along the three axes $z$, $y$, $x$ by respective angles $\gamma$, $\beta$, $\alpha$. $\mathbf{t} = (t_x ~ t_y ~ t_z)^\top$ is a $3 \times 1$ translation vector representing the position of the camera in $\{W\}$.

Let us now assume that we have a calibrated pinhole camera (i.e. we know its intrinsic parameters), which observes a set of 3D lines. Given $n \ge 9$ 3D lines $\mathbf{L}_i$ $(i = 1 \dots n)$ and their respective projections $\mathbf{l}_i$ onto the normalized image plane, we are able to estimate the camera pose. We parameterize the 3D lines using Pl\"{u}cker coordinates.

\subsection{Pl\"{u}cker coordinates of 3D lines}
\label{subsec:plucker}

3D lines can be represented using several parameterizations in the projective space \cite{bartoli2005structure}. Parameterization using Pl\"{u}cker coordinates is complete (i.e. every 3D line can be represented) but not minimal (a 3D line has 4 degrees of freedom but Pl\"{u}cker coordinate is a homogeneous 6-vector). The benefit of using Pl\"{u}cker coordinates is in convenient linear projection of 3D lines onto the image plane.

Given two distinct 3D points $\mathbf{A} = (a_x ~ a_y ~ a_z ~ a_w)^\top$ and $\mathbf{B} = (b_x ~ b_y ~ b_z ~ b_w)^\top$ in homogeneous coordinates, a line joining them can be represented using Pl\"{u}cker coordinates as a homogeneous 6-vector $\mathbf{L} = (\mathbf{u}^\top ~ \mathbf{v}^\top)^\top = (L_1 ~ L_2 ~ L_3 ~ L_4 ~ L_5 ~ L_6)^\top$, where

\vspace{-1em}
\begin{eqnarray}
	\label{eq:plucker}
	\mathbf{u}^\top &=& (L_1 ~ L_2 ~ L_3) = (a_x ~ a_y ~ a_z) ~ \times ~ (b_x ~ b_y ~ b_z) \\ \nonumber
	\mathbf{v}^\top &=& (L_4 ~ L_5 ~ L_6) = a_w(b_x ~ b_y ~ b_z) ~ - ~ b_w(a_x ~ a_y ~ a_z) \enspace,
\end{eqnarray}

\noindent '$\times$' denotes a vector cross product. The $\mathbf{v}$ part encodes direction of the line while the $\mathbf{u}$ part encodes position of the line in space. In fact, $\mathbf{u}$ is a normal of an interpretation plane -- a plane passing through the line and the origin. As a consequence, $\mathbf{L}$ must satisfy a bilinear constraint $\mathbf{u}^\top \mathbf{v} = 0$. Existence of this constraint explains the discrepancy between 4 degrees of freedom of a 3D line and its parameterization by a homogeneous 6-vector. More on Pl\"{u}cker coordinates can be found
in~\cite{hartley2004multiple}.

\vspace{-1em}
\subsection{Projection of 3D lines}
\label{subsec:projection}
\vspace{-0.3em}

3D lines can be transformed from the world coordinate system $\{W\}$ into the camera coordinate system $\{C\}$ using the $6 \times 6$ line motion matrix $\mathbf{T}$~\cite{bartoli20043d} as

\vspace{-0.5em}
\begin{equation}
  {\mathbf{L}}^{\scriptscriptstyle C} = \mathbf{T} {\mathbf{L}}^{\scriptscriptstyle W} \enspace .
  \label{eq:linetransform}
\end{equation}
\vspace{-1.7em}

\noindent The line motion matrix is defined as

\begin{figure}
\hspace{-2em}
\floatbox[
	\capbeside\thisfloatsetup{capbesideposition={right,center},capbesidewidth=0.55\linewidth}
]{figure}[0.85\FBwidth]
{
	\caption{
		3D line projection. The 3D line $\mathbf{L}$ is parameterized by its direction vector $\mathbf{v}$ and a normal $\mathbf{u}$ of its interpretation plane, which passes through the origin of the camera coordinate system $\{C\}$. Since the projected 2D line $\mathbf{l}$ lies at the intersection of the interpretation plane and the image plane, it is fully defined by the normal $\mathbf{u}$.
	}
	\label{fig:projection}
}
{
	\includegraphics[width=\linewidth]{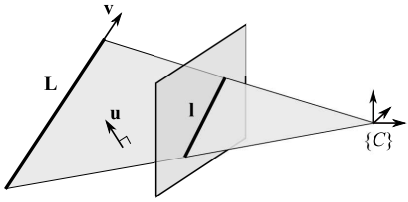}
	\hspace{-1.5em}
}
\vspace{-1.25em}
\end{figure}

\vspace{-0.5em}
\begin{equation}
  \mathbf{T} =
	\left( \begin{array}{lc}
		\mathbf{R} & \mathbf{R} [-\mathbf{t}]_{\times} \\
		\mathbf{0}_{\scriptscriptstyle 3 \times 3} & \mathbf{R} 
	\end{array} \right) \enspace ,
  \label{eq:linemotionmatrix}
\end{equation}

\noindent where $\mathbf{R}$ is a $3 \times 3$ rotation matrix and $[\mathbf{t}]_{\times}$ is a $3 \times 3$ skew-symmetric matrix constructed from the translation vector $\mathbf{t}$\footnote{Please note that our line motion matrix differs slightly from the matrix of Bartoli and Sturm~\cite[Eq.~(6)]{bartoli20043d}, namely in the upper right term: We have $\mathbf{R} [-\mathbf{t}]_{\times}$ instead of $[\mathbf{t}]_{\times} \mathbf{R}$ due to different coordinate system.}.
After 3D lines are transformed into the camera coordinate system, their projections onto the image plane can be determined as intersections of their interpretation planes with the image plane; see Figure~\ref{fig:projection} for illustration.

Recall from Eq.~(\ref{eq:plucker}) that coordinates of a 3D line consist of two 3-vectors: $\mathbf{u}$ (normal of an interpretation plane) and $\mathbf{v}$ (direction of a line). Since $\mathbf{v}$ is not needed to determine the projection of a line, only $\mathbf{u}$ needs to be computed. Thus, when transforming a 3D line according to Eq.~(\ref{eq:linetransform}) in order to calculate is projection, only the upper half of $\mathbf{T}$ is needed, yielding the $3 \times 6$ line projection matrix

\vspace{-0.8em}
\begin{equation}
  \mathbf{P} =
	\left( \begin{array}{ccc}
		\mathbf{R} & & \mathbf{R} [-\mathbf{t}]_{\times} 
	\end{array} \right) \enspace .
  \label{eq:lineprojectionmatrix}
\end{equation}
\vspace{-1.2em}

\noindent A 3D line ${\mathbf{L}}^{\scriptscriptstyle W}$ is then projected using the line projection matrix $\mathbf{P}$ as

\vspace{-0.5em}
\begin{equation}
  {\mathbf{l}}^{\scriptscriptstyle C} \approx \mathbf{P} {\mathbf{L}}^{\scriptscriptstyle W} \enspace ,
  \label{eq:lineprojection}
\end{equation}

\noindent where
${\mathbf{l}}^{\scriptscriptstyle C} = (l^{\scriptscriptstyle C}_x ~ l^{\scriptscriptstyle C}_y ~ l^{\scriptscriptstyle C}_w)^\top$
is a homogeneous 2D line in the normalized image plane and '$\approx$' denotes an equivalence of homogeneous coordinates, i.e. equality up to multiplication by a scale factor.

\vspace{-1em}
\subsection{Linear estimation of the line projection matrix}
\label{subsec:lineprojmat}
\vspace{-0.5em}

As the projection of 3D lines is defined by Eq.~(\ref{eq:lineprojection}), the problem of camera pose estimation resides in estimating the line projection matrix $\mathbf{P}$, which encodes all the six camera pose parameters $t_x$, $t_y$, $t_z$, $\alpha$, $\beta$, $\gamma$.

We solve this problem using the Direct Linear Transformation (DLT) algorithm, similarly to Hartley~\cite{hartley1998minimizing} who works with points. The system of linear equations (\ref{eq:lineprojection}) can be transformed into a homogeneous system

\vspace{-1.1em}
\begin{equation}
	\label{eq:system}
	\mathbf{M} \mathbf{p} = \mathbf{0}
\end{equation}

\noindent by transforming each equation of (\ref{eq:lineprojection}) so that only a 0 remains at the right-hand side. This forms a $2n \times 18$ measurement matrix $\mathbf{M}$ which contains coefficients of equations generated by correspondences between 3D lines and their projections $\mathbf{L}_i \leftrightarrow \mathbf{l}_i$ $(i = 1 \dots n, ~ n \ge 9)$. For details on construction of $\mathbf{M}$, please refer to Appendix~B in the supplementary material~\cite{pribyl2015supplementary}.

The DLT then solves (\ref{eq:system}) for $\mathbf{p}$ which is a 18-vector containing the entries of the line projection matrix $\mathbf{P}$.
Eq.~(\ref{eq:system}), however, holds only in the noise-free case. If a noise is present in the measurements, an inconsistent system is obtained.

\vspace{-0.5em}
\begin{equation}
	\label{eq:noisysystem}
	\mathbf{M} \mathbf{\hat{p}} = \boldsymbol{\epsilon}
\end{equation}

\noindent Only an approximate solution $\mathbf{\hat{p}}$ may be found through minimization of a $2n$-vector of measurement residuals $\boldsymbol{\epsilon}$ in the least squares sense on the right hand side of Eq.~(\ref{eq:noisysystem}).

Since DLT algorithm is sensitive to the choice of coordinate system, it is crucial to prenormalize the data to get properly conditioned $\mathbf{M}$~\cite{hartley1997defense}. Thanks to the principle of duality~\cite{coxeter2003projective}, coordinates of 2D lines can be treated as homogeneous coordinates of 2D points. The points should be translated and scaled so that their centroid is at the origin and their average distance from the origin is equal to $\sqrt{2}$.

The Pl\"{u}cker coordinates of 3D lines cannot be treated as homogeneous 5D points because of the bilinear constraint (see Section~\ref{subsec:plucker}). However, the closest point to a set of 3D lines can be computed using the Weiszfeld algorithm~\cite{aftab1lqclosest} and the lines can be translated so that the closest point is the origin.  

Once the system of linear equations given by (\ref{eq:noisysystem}) is solved in the least squares sense, \eg by Singular Value Decomposition (SVD) of $\mathbf{M}$, the estimate $\mathbf{\hat{P}}$ of the $3 \times 6$ line projection matrix $\mathbf{P}$ can be recovered from the 18-vector $\mathbf{\hat{p}}$.

\subsection{Estimation of the camera pose}
\label{subsec:estimcampose}

The $3 \times 6$ estimate $\mathbf{\hat{P}}$ of the line projection matrix $\mathbf{P}$ obtained as a least squares solution of Eq.~(\ref{eq:noisysystem}) does not satisfy the constraints imposed on $\mathbf{P}$. In fact, $\mathbf{P}$ has only 6 degrees of freedom -- the 6 camera pose parameters $t_x$, $t_y$, $t_z$, $\alpha$, $\beta$, $\gamma$. It has, however, 18 entries suggesting that it has 12 independent linear constraints, see Eq.~(\ref{eq:lineprojectionmatrix}). The first six constraints are imposed by the rotation matrix $\mathbf{R}$ that must satisfy the orthonormality constraints (unit-norm and mutually orthogonal rows). The other six constraints are imposed by the skew-symmetric matrix $[\mathbf{t}]_{\times}$ (three zeros on the main diagonal and antisymmetric off-diagonal elements). We propose the following method to extract the camera pose parameters from the estimate $\mathbf{\hat{P}}$.

First, the scale of $\mathbf{\hat{P}}$ has to be determined, since $\mathbf{\hat{p}}$ is usually of unit length as a minimizer of $\boldsymbol{\epsilon}$ in Eq.~(\ref{eq:noisysystem}). The correct scale of $\mathbf{\hat{P}}$ can be determined from its left $3 \times 3$ submatrix $\mathbf{\hat{P}}_1$ which is an estimate of the rotation matrix $\mathbf{R}$. Since the determinant of an orthonormal matrix must be equal to 1, $\mathbf{\hat{P}}$ has to be scaled by a factor $s = {1}/{\sqrt[3]{\det \mathbf{\hat{P}}_1}}$ so that $\det (s\mathbf{\hat{P}}_1) = 1$.

Second, the camera pose parameters can be extracted from $s\mathbf{\hat{P}}_2$, the scaled right $3 \times 3$ submatrix of $\mathbf{\hat{P}}$. The right submatrix is an estimate of a product of an orthonormal and a skew-symmetric matrix ($\mathbf{R}[-\mathbf{t}]_{\times}$) which has the same structure as the essential matrix~\cite{longuet1981computer} used in multi-view computer vision. Therefore, we use a method for the decomposition of an essential matrix into a rotation matrix and a skew-symmetric matrix (see~\cite[p. 258]{hartley2004multiple}) as follows: Let $s\mathbf{\hat{P}}_2$ = $\mathbf{U} \bm{\Sigma} \mathbf{V}^\top$ be the SVD of the scaled $3 \times 3$ submatrix $s\mathbf{\hat{P}}_2$, and let

\begin{equation}
  \mathbf{Z} = \left( \begin{array}{rrr}
		0 & 1 & 0 \\
		-1 & 0 & 0 \\
		0 & 0 & 0 
	\end{array} \right) \enspace , \enspace
	\mathbf{W} = \left( \begin{array}{rrr}
		0 & -1 & 0 \\
		1 & 0 & 0 \\
		0 & 0 & 1 
	\end{array} \right)\enspace .
  \label{eq:ZWmatrices}
\end{equation}

\noindent Two possible solutions (A and B) exist for the estimate $\mathbf{\hat{R}}$ of the rotation matrix and estimate $\hat{[\mathbf{t}]}_{\times}$ of the skew-symmetric matrix:

\begin{equation}
	\label{eq:solutions}
	\begin{array}{lcl}
		\mathbf{\hat{R}}_\mathrm{A} = \mathbf{UW} \;\; \mathrm{diag}(1 \; 1 \> \pm 1) \mathbf{V}^\top, & \hat{[\mathbf{t}]}_{\times\mathrm{A}} = \sigma \mathbf{VZ} \;\; \mathbf{V}^\top\\ 
		\mathbf{\hat{R}}_\mathrm{B} = \mathbf{UW}^\top \mathrm{diag}(1 \; 1 \> \pm 1) \mathbf{V}^\top, & \hat{[\mathbf{t}]}_{\times\mathrm{B}} = \sigma \mathbf{VZ}^\top \mathbf{V}^\top
	\end{array} \enspace ,
\end{equation}

\noindent where $\sigma = (\Sigma_{1,1} + \Sigma_{2,2}) / 2$ is an average of the first two singular values of $s\mathbf{\hat{P}}_2$ (a properly constrained essential matrix has the first and second singular values equal to each other and the third one is zero). The $\pm 1$ term in Eq.~(\ref{eq:solutions}) denotes either $1$ or $-1$ which has to be put on the diagonal so that $\det \mathbf{\hat{R}}_\mathrm{A} = \det \mathbf{\hat{R}}_\mathrm{B} = 1$.

The correct solution A or B is chosen based on a simple check whether 3D lines are in front of the camera or not. Extraction of the components $t_x$, $t_y$, $t_z$ of the translation vector from the skew symmetric matrix $[\mathbf{t}]_{\times}$ and also extraction of the rotation angles $\alpha$, $\beta$, $\gamma$ from the rotation matrix $\mathbf{R}$ are straightforward. This completes the pose estimation procedure.

Alternative ways of extracting the camera pose parameters from $s\mathbf{\hat{P}}$ also exist, \eg computing the closest rotation matrix $\mathbf{\hat{R}}$ to the left $3 \times 3$ submatrix of $s\mathbf{\hat{P}}_1$ and then computing  $\hat{[\mathbf{t}]}_{\times} = - \mathbf{\hat{R}}^\top s\mathbf{\hat{P}}_2$. However, our experiments showed that the alternative ways are less robust to image noise. Therefore, we have chosen the solution described in this section.

\vspace{0.5em} 
\subsection{Rejection of mismatched lines}
\label{subsec:outliers}

In practice, mismatches of lines (i.e. outlying correspondences) often occur, which degrades the performance of camera pose estimation. RANSAC algorithm is commonly used to identify and remove outliers; however, as our method works with 9 and more line correspondences, it is unsuitable for use in a RANSAC-like framework because the required number of correspondences leads to increased number of iterations.

For this reason, we use an alternative scheme called Algebraic Outlier Rejection (AOR) recently proposed by Ferraz \etal~\cite{ferraz2014very}. It is an iterative approach integrated directly into the pose estimation procedure (specifically, into solving Eq.~(\ref{eq:noisysystem}) in Section~\ref{subsec:lineprojmat}) in form of Iteratively Reweighted Least Squares. Wrong correspondences are identified as outlying based on the residual $\epsilon_i$ of the least squares solution in Eq.~(\ref{eq:noisysystem}). Correspondences with residuals above a predefined threshold $\epsilon_\mathrm{max}$ are assigned zero weights, which effectively removes them from processing in the next iteration, and the solution is recomputed. This is repeated until the error of the solution stops decreasing. %

The strategy for choosing $\epsilon_\mathrm{max}$ may be arbitrary but our experiments showed that the strategy $\epsilon_\mathrm{max} = \mathrm{Q}_{j}(\epsilon_1, \ldots, \epsilon_n)$ has a good tradeoff between robustness and the number of iterations. $\mathrm{Q}_j(\cdot)$ denotes the $j$th quantile, where $j$ decreases following the sequence (0.9, 0.8, $\ldots$ , 0.3) for the first 7 iterations and then it remains constant 0.25. This strategy usually leads to approximately 10 iterations.

It is important \emph{not} to prenormalize the data in this case because it will impede the identification of outliers. Prenormalization of inliers should be done just before the last iteration.

\section{Experimental evaluation}
\label{sec:evaluation}

Accuracy, robustness, and efficiency of the proposed algorithm were evaluated and compared with the state-of-the-art methods. The following methods were compared:

\newenvironment{tight_enumerate}{
	\begin{enumerate}
	\setlength{\itemsep}{0em}
	\setlength{\parskip}{0em}
}{\end{enumerate}}

\begin{tight_enumerate}
	\item \textbf{Mirzaei}, the method by Mirzaei and Roumeliotis~\cite{mirzaei2011globally} (results shown in red {\textcolor[rgb]{0.75,0,0}{\rule{0.5em}{0.5em}}}\,),
	\item \textbf{Zhang}, the method by Zhang \etal\cite{zhang2013robust} (results shown in blue {\textcolor[rgb]{0,0,0.75}{\rule{0.5em}{0.5em}}}\,),
	\item \textbf{ours}, the proposed method (results shown in green {\textcolor[rgb]{0,0.75,0}{\rule{0.5em}{0.5em}}}\,).
\end{tight_enumerate}

\noindent Both simulations using synthetic lines and experiments using  the real-world imagery are presented.

\vspace{-.75em}
\subsection{Synthetic lines}
\label{subsec:synthetic-lines}
\vspace{-.5em}

Monte Carlo simulations with synthetic lines were performed under the following setup: at each trial, $n$ 3D line segments were generated by randomly placing segment endpoints inside a cube $10^3$\,m large which was centered at the origin of $\{W\}$. A virtual pinhole camera with image size of $640 \times 480$\,pixels and focal length of 800\,pixels was placed randomly in the distance of 25\,m from the origin. The camera was then oriented so that it looked directly at the origin, having all 3D line segments in its field of view. The 3D line segments were projected onto the image plane. Coordinates of the 2D endpoints were then perturbed with independent and identically distributed Gaussian noise with standard deviation of $\sigma_\mathrm{p}$ pixels. 1000 trials were carried out for each combination of $n$, $\sigma_\mathrm{p}$ parameters.

Accuracy and robustness of each method was evaluated by measuring the estimated and true camera pose while varying $n$ and $\sigma_\mathrm{p}$ similarly to~\cite{mirzaei2011globally}. The position error $\Delta\tau = ||\mathbf{\hat{t}} - \mathbf{t}||$ is the distance from the estimated position $\mathbf{\hat{t}}$ to the true position $\mathbf{t}$. The orientation error $\Delta\Theta$  was calculated as follows. The difference between the true and estimated rotation matrix ($\mathbf{R}^\top \mathbf{\hat{R}}$) is converted to axis-angle representation ($\mathbf{e}$, $\theta$) and the absolute value of the difference angle $|\theta|$ is considered as the orientation error.

\begin{figure}[h]
\centering
\includegraphics[width=0.99\linewidth]{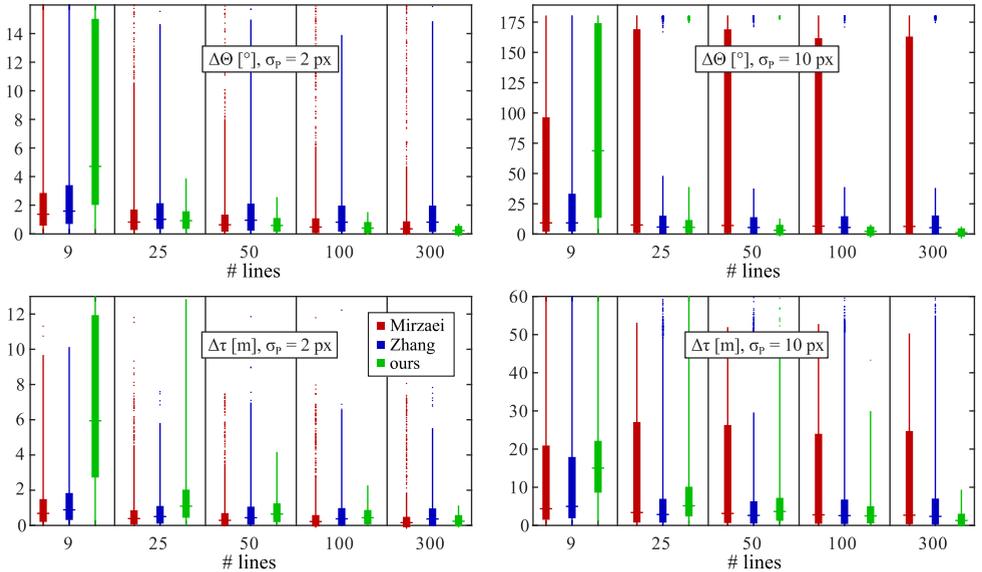}
\caption{
The distribution of orientation errors ($\Delta\Theta$, \textbf{top}) and  position errors ($\Delta\tau$, \textbf{bottom}) in estimated camera pose as a function of the number of lines. Two levels of Gaussian noise are depicted: with standard deviation of $\sigma_\mathrm{p}=2$\,px (\textbf{left}) and with $\sigma_\mathrm{p}=10$\,px (\textbf{right}).
Each box depicts the median (\emph{dash}), interquartile range - IQR (\emph{box body}), minima and maxima in the interval of $10 \times$ IQR (\emph{whiskers}) and outliers (\emph{isolated dots}).
}
\label{fig:errors}
\end{figure}

As illustrated in Figure~\ref{fig:errors}, 25 lines are generally enough for our method to be on par with the state-of-the-art in terms of accuracy. 50 and more lines are usually exploited better by our method. As the number of lines grows, our method becomes even more accurate than the others. It should be noted that  the orientation error decreases more rapidly than the position error with the number of lines. Our method is outperformed by the others in the minimal case of 9 lines. However, as soon as more lines are available, the results of our approach rapidly improve. This fact is a matter of chosen parameterization. Pl\"{u}cker coordinates of 9 lines are just enough to define all 18 entries of the line projection matrix $\mathbf{P}$ in Eq.~(\ref{eq:lineprojectionmatrix}). More lines bring redundancy into the system and compensate for noise in the measurements. However, even 9 lines are enough to produce an exact solution in a noise-free case.

All the three methods sometimes yield an improper estimate with exactly opposite orientation. This can be observed as isolated dots particularly in Figure~\ref{fig:errors} (top, right). Furthermore, the method of Mirzaei sometimes produced an estimate where the camera is located in between the 3D lines and it has random orientation. This happened more frequently in the presence of stronger image noise, as it is apparent from increased red bars in Figure~\ref{fig:errors} (right). The robustness of Mirzaei's method is thus much lower compared to our method and Zhang's method. However, the method of Zhang sometimes produced a degenerate pose estimate very far from the correct camera position when the 3D lines projected onto a single image point (this phenomenon cannot be seen in Figure~\ref{fig:errors} as such estimates are out of scale of the plots). The proposed method does not suffer from any of these two issues and is more robust in cases with 50 and more lines.

\subsection{Real images}

The three methods were also tested using real-world images from the VGG Multiview Data\-set\footnote{\url{http://www.robots.ox.ac.uk/\textasciitilde vgg/data/data-mview.html}}. It contains indoor and outdoor image sequences of buildings with extracted 2D line segments, their reconstructed positions in 3D, and camera projection matrices. Each method was run on the data and the estimated camera poses were used to reproject the 3D lines onto the images to validate the results.

\begin{figure}[h]
	\centering
	\includegraphics[width=\linewidth]{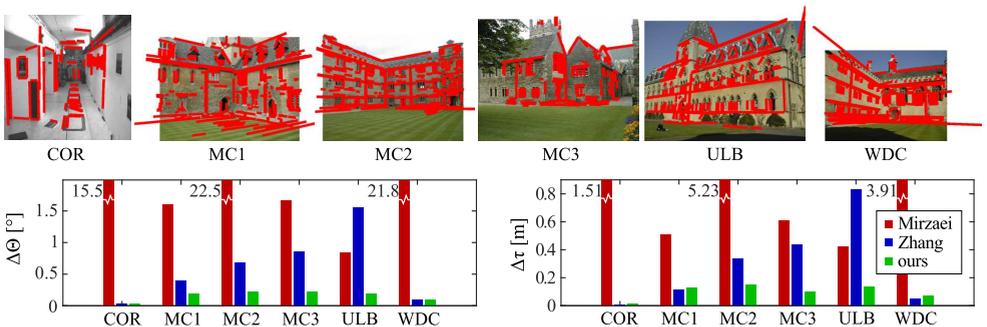}
	\caption{
	(\textbf{top}) Example images from the VGG dataset overlaid with reprojections of 3D line segments using our estimated camera pose. (\textbf{bottom}) Average camera orientation error $\Delta\Theta = |\theta|$ and average position error $\Delta\tau = ||\mathbf{\hat{t}} - \mathbf{t}||$ in individual image sequences.
	}
	\label{fig:VGG_dataset}
	\vspace{-1em}
\end{figure}

The proposed algorithm performs similarly or better than Zhang's method while Mirzaei's method behaves noticeably worse, as it can be seen in Figure~\ref{fig:VGG_dataset} and Table~\ref{table:results}. Detailed results with all images from the sequences are available as supplementary material~\cite{pribyl2015supplementary}.

\setlength{\tabcolsep}{3pt}
\begin{table}
	\begin{center}
		{\fontsize{.95em}{1.2em}\selectfont
		\begin{tabular}{lrrccccccccc}
			\hline\noalign{\smallskip}
			 & \#\hspace{0.5em} & \#\hspace{0.8em} & & \multicolumn{2}{c}{\textbf{Mirzaei}} & & \multicolumn{2}{c}{\textbf{Zhang}} & & \multicolumn{2}{c}{\textbf{ours}}\\
			Sequence & lines & imgs. & & $\Delta\Theta$ & $\Delta\tau$ & & $\Delta\Theta$ & $\Delta\tau$ & & $\Delta\Theta$ & $\Delta\tau$\\
			\noalign{\smallskip}
			\hline
			\noalign{\smallskip}
			Corridor           &  69 & 11 & & 15.510\,$^{\circ}$ & 1.510\,m & & \textbf{0.029\,$^{\circ}$} & \textbf{0.008\,m} & & 0.034\,$^{\circ}$ & 0.013\,m\\
			Merton College I   & 295 &  3 & & \hspace{0.5em}1.610\,$^{\circ}$ & 0.511\,m & & 0.401\,$^{\circ}$ & \textbf{0.115\,m} & & \textbf{0.195\,$^{\circ}$} & 0.128\,m\\
			Merton College II  & 302 &  3 & & 22.477\,$^{\circ}$ & 5.234\,m & & 0.676\,$^{\circ}$ & 0.336\,m & & \textbf{0.218\,$^{\circ}$} & \textbf{0.151\,m}\\
			Merton College III & 177 &  3 & & \hspace{0.5em}1.667\,$^{\circ}$ & 0.608\,m & & 0.859\,$^{\circ}$ & 0.436\,m & & \textbf{0.223\,$^{\circ}$} & \textbf{0.101\,m}\\
			University Library & 253 &  3 & & \hspace{0.5em}0.837\,$^{\circ}$ & 0.423\,m & & 1.558\,$^{\circ}$ & 0.833\,m & & \textbf{0.189\,$^{\circ}$} & \textbf{0.138\,m}\\
			Wadham College     & 380 &  5 & & 21.778\,$^{\circ}$ & 3.907\,m & & 0.103\,$^{\circ}$ & \textbf{0.047\,m} & & \textbf{0.086\,$^{\circ}$} & 0.072\,m\\
			\hline
		\end{tabular}
		
		}
		\vspace{-1.5em}
	\end{center}
	\caption{Results of the methods on the VGG dataset in terms of average camera orientation error $\Delta\Theta = |\theta|$ and average position error $\Delta\tau = ||\mathbf{\hat{t}} - \mathbf{t}||$. Best results are in bold.}
	\label{table:results}
\end{table}
\setlength{\tabcolsep}{1.4pt}

\subsection{Efficiency}

\begin{figure}[h]
\begin{floatrow}
\capbtabbox{
	\setlength{\tabcolsep}{4pt}
	{
		\fontsize{0.95em}{1.2em}\selectfont
		\begin{tabular}{lrrrr}
		\hline\noalign{\smallskip}
		\#\,lines & 9 & 100 & 1000\\
		\noalign{\smallskip}
		\hline
		\noalign{\smallskip}
		Mirzaei  & 72.0 & 79.5 & 168.2\\
		Zhang & 8.7 & 42.1 & 899.4\\
		ours & \textbf{3.2} & \textbf{3.8} & \textbf{28.5}\\
		\hline
		\end{tabular}
	}
	\setlength{\tabcolsep}{1.4pt}
	\vspace{2em}
}{
	\caption{
		Runtimes in milliseconds for varying number of lines, averaged over 1000 runs.
	}
	\label{table:time}
}
\ffigbox[21em]{
	\includegraphics[width=\linewidth]{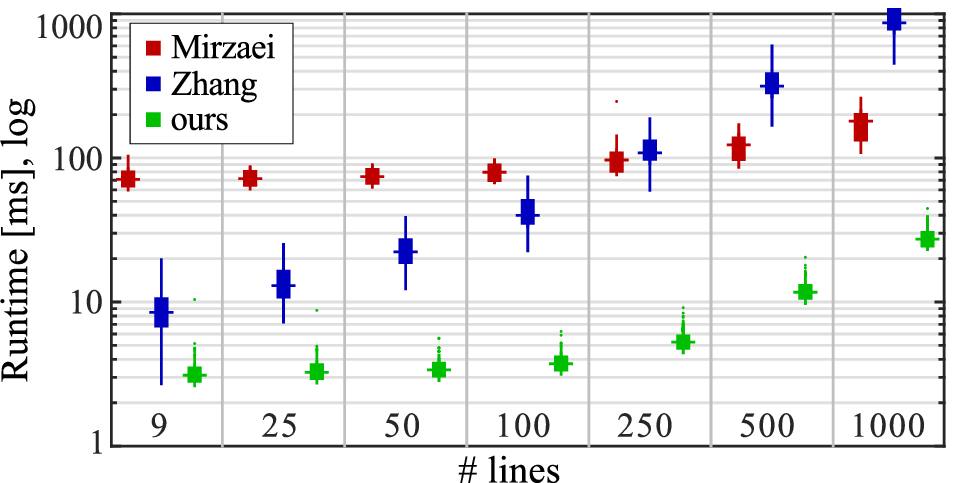}
	\vspace{-2em}
}{
	\caption{
		The distribution of runtimes as a function of the number of lines. Logarithmic vertical axis.
		Meaning of the boxes is the same as in Figure~\ref{fig:errors}.
	}
	\label{fig:time}
}
\end{floatrow}
\vspace{-.5em}
\end{figure}

Efficiency of each method was evaluated by measuring runtime on a desktop PC with a quad core Intel i5 3.33\,GHz CPU. Matlab implementations downloaded from the websites of the respective authors were used. As it can be seen in Table~\ref{table:time} and Figure~\ref{fig:time}, our method significantly outperforms the others in terms of speed. Computational complexity of all evaluated methods is linearly dependent on the number of lines. However, the absolute numbers differ substantially.
Mirzaei's method is slower than Zhang's method for up to cca 200 lines. This is due to computation of a $120 \times 120$ Macaulay matrix in Mirzaei's method which has an effect of a constant time penalty. However, Zhang's method is slower than Mirzaei's for more than 200 lines. Our method is the fastest no matter how many lines are processed; it is approximately one order of magnitude faster than both competing methods. The linear computational complexity of our method is only achieved due to the prenormalization of input data and subsequent SVD of the $2n \times 18$ measurement matrix $\mathbf{M}$; all the other computations are performed in constant time.

\vspace{-1em}
\subsection{Robustness to outliers}
\vspace{-.5em}

As a practical requirement, robustness to outlying correspondences was also tested. The experimental setup was the same as in Section~\ref{subsec:synthetic-lines}, using $n = 500$\,lines which endpoints were perturbed with slight image noise with $\sigma_{\mathrm{p}} = 2$\,pixels. The image lines simulating outlying correspondences were perturbed with an aditional extreme noise with $\sigma_{\mathrm{p}} = 100$\,pixels.
The methods of Mirzaei and Zhang were plugged into a MLESAC (an improved version of RANSAC)~\cite{torr2000mlesac} framework, generating camera pose hypotheses from 3 and 4 randomly selected line correspondences, respectively. The inlying correspondences were identified based on the line reprojection error~\cite{taylor1995structure}. No heuristics for early hypothesis rejection was utilized, as it can also be incorporated into the Algebraic Outlier Rejection scheme, \eg by weighting the line correspondences. The proposed method with AOR was set up as described in Section~\ref{subsec:outliers}.

\begin{figure}[h]
	\includegraphics[width=0.8\linewidth]{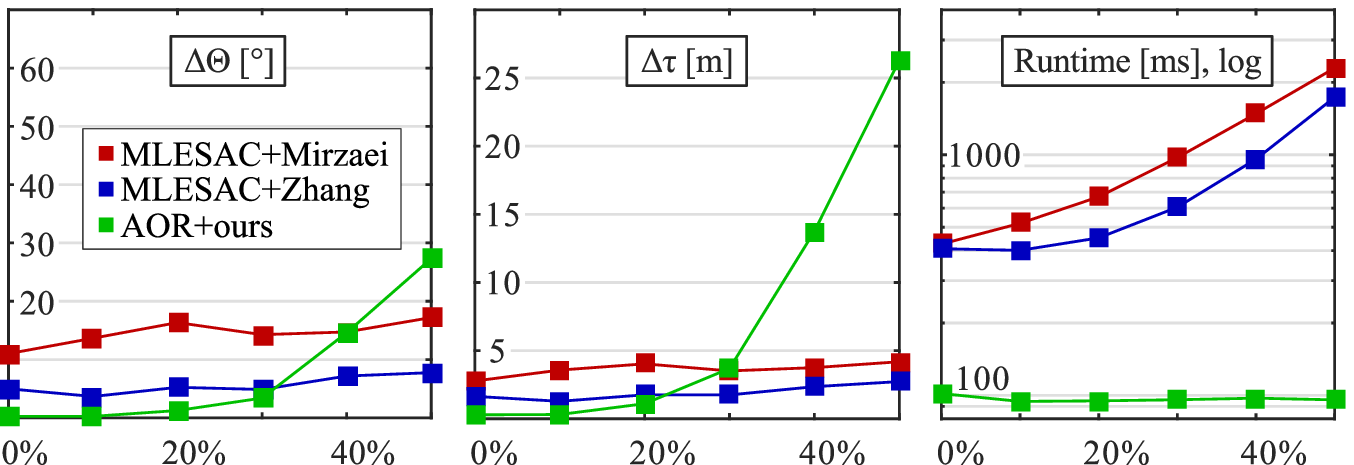}
	\caption{
		Camera pose errors (\textbf{left}, \textbf{center}) and runtime (\textbf{right}) depending on the percentage of outliers. $n = 500$\,lines, $\sigma_{\mathrm{p}} = 2$\,pixels, averaged over 1000 runs.
	}
	\label{fig:outliers}
	\vspace{-0.5em}
\end{figure}

While the RANSAC-based approaches can theoretically handle any percentage of outliers, the proposed method with AOR has a break-down point at about 30\,\% of outliers, as depicted in Figure~\ref{fig:outliers}. However, for the lower percentage of outliers, our method is more accurate and 5-7$\times$ faster.

The original AOR approach applied to the PnP problem~\cite{ferraz2014very} has a higher break-down point at 45\,\%. We think it might be because the authors need to estimate a null space with only 12 entries whereas we estimate 18 entries of the nullspace $\mathbf{\hat{p}}$ in Eq.~(\ref{eq:noisysystem}). The use of barycentric coordinates for parameterization of 3D points in~\cite{ferraz2014very} may also play a role.

\section{Conclusions}
\label{sec:conclusions}
\vspace{-0.3em}

In this paper, a novel algebraic approach to the Perspective-n-Line problem is proposed. The approach is substantially faster, yet equally accurate and robust compared to the state-of-the-art. The superior computational efficiency of the proposed method achieving speed-ups of more than one order of magnitude for high number of lines is proved by simulations and experiments. As an alternative to the commonly used RANSAC, Algebraic Outlier Rejection is used to deal with mismatched lines. The proposed method requires at least 9 lines, but it is particularly suitable for large scale and noisy scenarios. For very small size noisy scenarios ($\leq 25$ lines), the state-of-the-art performs better and we recommend to use the Zhang's method. Future work involves examination of the degenerate line configurations.

The Matlab code of the proposed method and the appendices are publicly available in the supplementary material~\cite{pribyl2015supplementary}.

\vspace{-0.5em}
\paragraph{Acknowledgements}
This work was supported by the Technology Agency of the Czech Republic by projects TA02030835 D-NOTAM and TE01020415 V3C. It was also supported by SoMoPro II grant (financial contribution from the EU 7 FP People Programme Marie Curie Actions, REA 291782, and from the South Moravian Region). The content of this article does not reflect the official opinion of the European Union. Responsibility for the information and views expressed therein lies entirely with the authors.

\bibliography{bmvc2015_id125}
\end{document}